\documentclass[11pt]{article}
\usepackage[final]{acl}
\usepackage{times}
\usepackage{latexsym}
\usepackage{amsmath}
\usepackage{amssymb}
\usepackage{graphicx}
\usepackage{booktabs}
\usepackage{multirow}
\usepackage{xcolor}
\usepackage{url}
\usepackage{subcaption}
\usepackage{colortbl}

\title{Metric-Dependent Annotation Saturation\\for Learning from Label Distributions}

\author{Guneet Kohli \\
  Apple  \\   
  \texttt{g\_kohli@apple.com}}   

\begin{document}
\maketitle

\begin{abstract}
When annotators disagree on a label, the disagreement itself
carries signal---and the number of annotators needed to capture
it depends on the evaluation metric.
We fine-tune NLI models on label distributions subsampled from
ChaosNLI, a dataset providing 100 independent annotator judgments
per item, and identify \emph{metric-dependent saturation}.
In our 3-class NLI setting, entropy correlation---whether the model
identifies \emph{which} items elicit disagreement---requires
$N \approx 20$--$50$ annotators to converge, while distributional
match (KL divergence) saturates by $N \approx 10$
(87--95\% of improvement across five model seeds); the specific
thresholds are estimates for this setting, and it is the phenomenon,
not the constants, that we expect to generalize.
This finding rests on a prior observation:
soft labels carry item-specific signal that label smoothing cannot
replicate. Across five smoothing intensities, entropy
correlation clusters at $r \approx 0.45$--$0.49$,
while soft labels reach $r = 0.643$ ($p < 0.001$); per-item
analysis traces this gap to smoothing's inability to distinguish
ambiguous items from clear ones.
The soft-label advantage replicates across two architectures
(DeBERTa, RoBERTa) and a non-NLI-pretrained baseline;
differential saturation additionally appears in an
exploratory cross-domain evaluation on content safety.
These results suggest that annotation budgets should be informed
by the target evaluation metric rather than set uniformly.
\end{abstract}

\section{Introduction}
\label{sec:intro}

Natural language inference (NLI) is a foundational task in NLP, yet the standard
practice of training on majority-vote labels discards the vast majority of
annotation signal. When 100 annotators label an NLI item, a typical split might
be 62\% entailment, 25\% neutral, 13\% contradiction---yet the model receives
only ``entailment.'' This collapses a rich distribution into a point estimate,
producing models that are systematically overconfident on genuinely ambiguous items
\citep{nie2020what, pavlick2019inherent}.

Training on the full annotator distribution---so-called \emph{soft-label}
training---has shown promise in computer vision \citep{peterson2019human} and
is increasingly studied for NLP \citep{uma2021learning, fornaciari2021beyond}.
However, a natural skepticism persists: \emph{label smoothing}
\citep{szegedy2016rethinking}, which blends one-hot labels with a uniform
distribution, produces soft targets at zero annotation cost. One might expect
that on ambiguous items---where humans genuinely disagree---the uniform
distribution is a reasonable approximation to the true label distribution.
If so, a well-tuned smoothing parameter $\alpha$ could match 100 annotators
without additional annotation cost, rendering the expensive collection effort redundant.

We test this hypothesis directly and find that it fails except at high
smoothing intensity, and then only where the human distribution
approaches uniformity.
Moderate smoothing ($\alpha = 0.3$) yields nearly constant error across
entropy terciles: it neither adapts to ambiguous items nor spares clear
ones. Only heavy smoothing ($\alpha = 0.5$) approximates the target on
the most ambiguous items, and the same redistribution that helps there
more than doubles the error on clear items.
This pattern---invisible in aggregate metrics---reveals the fundamental
limitation of \emph{item-agnostic} uncertainty: a model cannot learn
\emph{which} items are ambiguous by being told that \emph{all} items are
ambiguous.

Having established that soft labels carry irreplaceable item-specific
signal, the natural question is whether this signal can be obtained
efficiently: how many annotators per item are needed to capture it?
We construct annotation efficiency curves by subsampling $N \in \{3, 5,
10, 20, 50, 100\}$ annotators and observe
\emph{metric-dependent saturation} (which we also call
\emph{differential saturation}):
different evaluation metrics plateau at different annotator counts.
Distributional match (KL divergence) saturates rapidly because it is
dominated by majority-class probability, which few annotators estimate well.
Uncertainty ranking (entropy correlation) improves more slowly because it
depends on the full distributional shape, including minority-class
proportions that require more annotators to resolve.

Our contributions are:
\begin{enumerate}
\item \textbf{Metric-dependent annotation saturation.}
  KL divergence to human distributions saturates by $N \approx 10$
  annotators (87--95\% of hard$\rightarrow$soft improvement across
  five model seeds), while
  entropy correlation---the rank agreement between model uncertainty
  and human disagreement---lags behind, requiring $N \approx 20$--$50$
  to converge.
  Annotation budgets should be set according to the target
  evaluation metric.

\item \textbf{Soft labels are qualitatively separated from label smoothing.}
  Across five label-smoothing intensities
  ($\alpha \in \{0.05, 0.1, 0.2, 0.3, 0.5\}$),
  entropy correlation clusters at
  $r \in [0.446, 0.489]$, while soft labels reach $r = 0.643$
  ($\Delta = +0.154$ on DeBERTa; raw $p < 0.001$, significant under
  Holm-Bonferroni correction).
  The gap replicates on RoBERTa-large ($\Delta = +0.249$).

\item \textbf{Per-item diagnosis of when smoothing fails.}
  Stratifying by human disagreement, label smoothing at $\alpha = 0.3$
  is nearly constant across entropy terciles (KL $\approx$ 0.177),
  never beating soft labels on any tercile, including high-entropy
  items (0.177 vs.\ 0.167, n.s.). Only heavy smoothing ($\alpha = 0.5$) edges out
  soft labels on the most ambiguous tercile (0.143 vs.\ 0.167), while
  degrading clear items to 0.265 vs.\ 0.121. Item-agnostic smoothing
  thus captures ambiguity only by sacrificing clear items;
  item-specific annotations avoid this trade-off entirely.
\end{enumerate}

\section{Related Work}
\label{sec:related}

\paragraph{Soft-label training lacks a label-smoothing control.}
\citet{peterson2019human} demonstrated that training image classifiers on
soft labels from ${\sim}51$ annotators per image (CIFAR-10H) improves both
robustness and calibration, and subsequent work has explored NLI
distribution training \citep{meissner2021embracing}, multi-task
variants \citep{fornaciari2021beyond, davani2022dealing}, efficient
elicitation \citep{collins2022eliciting}, and integration with
single-label classifiers
\citep{wu2023dont}. \citet{uma2021learning} survey the broader landscape of
learning from annotator disagreement.
These studies do not compare soft-label training against a systematic
\emph{label-smoothing sweep} as a zero-cost control: without showing
where smoothing fails, the case for expensive soft-label collection
remains incomplete. We provide this control.

\paragraph{Systematic annotation efficiency curves are absent from the literature.}
The crowdsourcing literature has long studied cost--quality tradeoffs
for label acquisition \citep{sheng2008get}, but that work focuses on
noise reduction for majority-vote accuracy rather than distributional
fidelity.
\citet{zhang2021capturing} compare annotation budgets with a coarse-grained
comparison (10K$\times$1 vs.\ 1K$\times$10 annotators) but do not
construct fine-grained efficiency curves that reveal saturation
behavior.
\citet{zhou2022distributed} train models to predict human opinion
distributions using Monte Carlo Dropout and ensembles on ChaosNLI
\citep{nie2020what}, but do not study how many annotators
are needed.
\citet{kurniawan2025training} compare 14 training methods across six datasets
but do not study subsampling or calibration.
\citet{singh2025softlabel} train on ChaosNLI soft labels using frozen
embeddings with an MLP head, reporting entropy correlation of 0.284---but
do not vary annotator count or compare against label smoothing.
To our knowledge, we provide the first systematic efficiency curve
for NLI, revealing that different metrics saturate at different rates.

\paragraph{Can calibration substitute for annotation? The answer is metric-dependent.}
Label smoothing \citep{szegedy2016rethinking} replaces one-hot targets with
$(1-\alpha) \cdot \mathbf{e}_y + \alpha/K$, applying \emph{uniform}
uncertainty to all items.
\citet{muller2019when} show that label smoothing improves calibration by
preventing overconfident predictions, and \citet{xia2025understanding}
provide a gradient-level explanation for why it degrades
\emph{selective} classification: smoothing regularizes the max logit
more for correct predictions, degrading the uncertainty rank ordering.
Closest to our question, \citet{wang2022capture} show that calibration
techniques---label smoothing and temperature scaling---let NLI models
capture human disagreement distributions without collecting ambiguity
annotations, evaluating primarily with global divergence measures.
Our results delimit when that conclusion holds: item-agnostic
calibration recovers \emph{distribution match}---the metric family
that saturates with few annotators---but no smoothing intensity
recovers \emph{uncertainty ranking}, precisely the signal that
requires many. Wang et al.\ do not vary annotator counts; the
saturation analysis and the smoothing sweep across intensities and
budgets are, to our knowledge, new here.

\paragraph{Calibration metrics need updating for distributional targets.}
\citet{guo2017calibration} introduced Expected Calibration Error (ECE) and
\citet{desai2020calibration} showed that fine-tuned transformers become
overconfident. But standard ECE evaluates against majority-vote labels:
a model that correctly outputs $P(\text{entailment}) = 0.62$ when humans
split 62/25/13 is \emph{penalized} as ``underconfident.''
\citet{baan2022stop} make this critique precise, showing that
calibration to the majority class is theoretically problematic under
inherent disagreement and deriving instance-level alternatives on
ChaosNLI.
\citet{khurana2024crowd} explore whether annotator disagreement can
inform calibration in subjective tasks, but do not study how annotation
density affects calibration metrics.
We evaluate with distribution-aware metrics---Brier score against the
human distribution and distributional ECE---following the proper scoring
rule framework of \citet{gneiting2007strictly}.

\paragraph{Modeling disagreement is advocated in principle but under-studied in practice.}
A growing body of work argues that human label variation is signal, not
noise \citep{plank2022problem, pavlick2019inherent, basile2021we},
a perspectivist position now institutionalized in the LeWiDi shared
tasks on learning with disagreements \citep{leonardelli2023semeval},
whose recent edition features label-distribution-learning entries
closest in spirit to our training setup \citep{ignatev2025demeva},
and Bayesian annotation models can estimate annotator reliability and
item difficulty from crowdsourced labels
\citep{dawid1979maximum, paun2018comparing}.
\citet{weerasooriya2023crowdopinion} predict label distributions by
clustering similar items, evaluating with KL divergence---but do not
study how the number of annotators per item affects downstream model
training. These papers focus on \emph{whether} and \emph{how} to model
disagreement rather than \emph{how much} annotation density is needed
to do so effectively. We provide the empirical bridge: systematic
evidence for how annotation budgets interact with evaluation metrics
when training on distributions.

\section{Method}
\label{sec:method}

\subsection{Task and Data}
\label{sec:data}

We use ChaosNLI \citep{nie2020what}, combining ChaosNLI-MNLI (1,599 items
from MultiNLI; \citealt{williams2018broad}) and ChaosNLI-SNLI (1,514 items
from SNLI; \citealt{bowman2015large}), for a total of 3,113 items, each
with 100 independent annotations across three classes: entailment, neutral,
and contradiction. For each item $i$, the human label distribution is
$\mathbf{h}_i = [h_i^e, h_i^n, h_i^c]$, where $h_i^k$ is the fraction
of annotators who selected class $k$.

We split the data into 2,179 training, 467 validation, and 467 test items
(70/15/15, stratified by plurality label). All results report test-set
performance. We verify robustness across three data splits
(split seeds 42, 43, 44).

Very few NLP datasets provide $\geq$50 annotators per item;
most benchmarks collect 3--5. This density is a prerequisite for
constructing meaningful saturation curves---one cannot study whether
$N{=}20$ suffices without ground truth from $N{=}100$.
Our cross-domain evaluation uses DICES-990 \citep{aroyo2024dices},
which provides ${\sim}70$ annotators per item on a content safety task
(\S\ref{sec:dices_main}).

\subsection{Training Configurations}
\label{sec:configs}

All configurations fine-tune DeBERTa-v3-base \citep{he2023debertav3},
initialized from a checkpoint pre-fine-tuned on MNLI, FEVER, and ANLI.
We compare seven training configurations (plus post-hoc temperature
scaling applied to all):

\paragraph{Hard-label.} Standard cross-entropy against the majority-vote
label $\mathbf{e}_y$ ($K=3$). For one-hot targets, KL divergence reduces
to cross-entropy (differing only by the constant $H(\mathbf{e}_y) = 0$),
so this baseline is mathematically equivalent to standard cross-entropy
training.

\paragraph{Hard-label + temperature scaling.} Post-hoc calibration
\citep{guo2017calibration} applied to the hard-label model. A single
temperature $T > 0$ is optimized on the validation set; at test time,
$\mathbf{p} = \text{softmax}(\mathbf{z}/T)$.

\paragraph{Label smoothing (LS; $\alpha \in \{0.05, 0.1, 0.2, 0.3, 0.5\}$).}
Targets are constructed as:
\begin{equation}
t_{i,k}^\text{LS} = (1-\alpha) \cdot \mathbf{1}[k=y] + \frac{\alpha}{K}\,.
\label{eq:ls}
\end{equation}
We train with KL divergence loss $\mathcal{L} = \text{KL}(\mathbf{t}_i^\text{LS} \| \mathbf{p}_\theta(x_i))$.
This provides a zero-cost control: if any $\alpha$ matches soft labels,
then 100 annotators add nothing beyond a free hyperparameter.

\paragraph{Soft-label.} Targets are the full 100-annotator distribution
$\mathbf{h}_i$. We train with KL divergence:
\begin{equation}
\mathcal{L}_\text{soft} = \text{KL}(\mathbf{h}_i \| \mathbf{p}_\theta(x_i))
  = \sum_{k=1}^{K} h_i^k \log \frac{h_i^k}{p_\theta^k(x_i)}\,.
\label{eq:soft}
\end{equation}

\subsection{Evaluation Metrics}
\label{sec:metrics}

We distinguish two families of metrics based on the evaluation reference.

\paragraph{Majority-vote metrics.} \textbf{Accuracy}: whether the model's
top prediction matches the plurality label. \textbf{ECE}
\citep{guo2017calibration}: expected calibration error against majority labels.
These metrics treat the majority vote as ground truth and
penalize models that express calibrated uncertainty.

\paragraph{Distribution-aware metrics.}
These compare the model's predicted distribution $\mathbf{p}_\theta(x_i)$
against the human distribution $\mathbf{h}_i$:

\textbf{Brier-soft} \citep{brier1950verification}: the mean squared error
between model and human distributions,
$\frac{1}{n} \sum_i \sum_k (p_\theta^k(x_i) - h_i^k)^2$,
a strictly proper scoring rule.

\textbf{KL divergence}: $\text{KL}(\mathbf{h}_i \| \mathbf{p}_\theta(x_i))$,
measuring distributional alignment. Note that this metric is identical to
the soft-label training loss (Eq.~\ref{eq:soft}) evaluated on held-out
items, so the soft-label model directly optimizes evaluation KL while
hard-label and LS models do not. This is inherent to the comparison---not
circularity, since training and evaluation items are disjoint---but it
motivates entropy correlation and Brier-soft as complementary metrics
that are not aligned with any model's training objective.

\textbf{Dist-ECE}: Per-class calibration error---for each class $k$,
bins items by $p_\theta^k(x_i)$ and measures the gap to the average
human $h_i^k$ in each bin, averaged across classes.

\textbf{Entropy correlation} (correlation \emph{of} entropies, not an
information-theoretic quantity): Pearson $r$ between model prediction
entropy $H(\mathbf{p}_\theta(x_i)) = -\sum_k p_\theta^k \log_2 p_\theta^k$
and human label entropy $H(\mathbf{h}_i)$.%
\footnote{Spearman $\rho$ yields near-identical conclusions throughout
(e.g., soft $\rho{=}0.611$ vs.\ LS range $[0.442, 0.467]$ vs.\ hard $0.466$;
cf.\ Pearson $r{=}0.643$ vs.\ $[0.446, 0.489]$ vs.\ $0.487$).
We report Pearson $r$ in the main text as the relationship is
approximately linear (Appendix~\ref{app:spearman}).}

We highlight entropy correlation because it captures whether the model
has learned \emph{which} items are ambiguous, independent of absolute
probability values---a property relevant to selective prediction and
active learning.
However, our central argument is that different metrics
answer different questions, so no single metric should be privileged
universally.

\paragraph{Guarding against evaluation circularity.}
Because the soft-label model trains on the same \emph{kind} of signal
it is evaluated against, three design choices jointly rule out
circular evaluation.
(1)~\emph{Held-out annotators}: retraining soft-label models on
distributions built from 80 of the 100 annotators and evaluating every
configuration against the remaining 20 (\S\ref{sec:analysis})
preserves all conclusions on annotation data no model ever saw.
(2)~\emph{Non-objective metrics}: the headline separation result is on
entropy correlation (and Brier-soft), which is not the training
objective of any configuration.
(3)~\emph{Disjoint items}: train and test items never overlap, so even
the objective-aligned KL metric is measured on unseen items.

\subsection{Brier Score Decomposition}
\label{sec:decomposition}

To understand \emph{why} soft labels outperform label smoothing, we extend
the classical Brier score decomposition \citep{murphy1973new} to
distributional ground truth. For each class $k$, we bin items by the
model's predicted $p_\theta^k$ and decompose:
\begin{equation}
\text{BS}_\text{soft} = \underbrace{\text{REL}}_{\text{calibration}}
  - \underbrace{\text{RES}}_{\text{resolution}}
  + \underbrace{\text{UNC}}_{\text{irreducible}}
\label{eq:murphy}
\end{equation}
where, summing over classes and bins:
\begin{align}
\text{REL} &= \frac{1}{n} \sum_k \sum_j n_j \left(\bar{p}_j^k - \bar{h}_j^k\right)^2\,,
\label{eq:rel} \\
\text{RES} &= \frac{1}{n} \sum_k \sum_j n_j \left(\bar{h}_j^k - \bar{h}^k\right)^2\,,
\label{eq:res} \\
\text{UNC} &= \frac{1}{n} \sum_k \sum_i \left(h_i^k - \bar{h}^k\right)^2\,.
\label{eq:unc}
\end{align}
Here $\bar{p}_j^k$ and $\bar{h}_j^k$ are the mean model and human
probabilities for class $k$ in bin $j$, $n_j$ is the bin count, and
$\bar{h}^k$ is the global mean human probability for class $k$.

\textbf{Resolution} (RES) directly measures what entropy correlation
captures qualitatively: can the model's confidence bins separate items
with genuinely different human disagreement patterns?
A model that assigns the same uncertainty to every item (like label
smoothing) has RES $\approx 0$; a model that accurately sorts items
by ambiguity has high RES.

\subsection{Annotation Efficiency Curve}
\label{sec:efficiency}

To determine how many annotators are needed, we subsample
$N \in \{3, 5, 10, 20, 50, 100\}$ annotators from the full 100 per item
and construct soft labels from the subsample. For each $N$, we train five model seeds (42--46),
each with five independent subsampling seeds, yielding 25
observations per $N$.

The hard-label baseline uses the same fixed model seed with majority-vote
targets. We report each metric as the percentage of hard$\rightarrow$soft(100)
improvement captured at each $N$.

\subsection{Statistical Analysis}

All comparisons use five random seeds for model initialization. We report
means and standard deviations across seeds, with paired $t$-tests corrected
for multiple comparisons via Holm-Bonferroni. We report absolute deltas as
the primary effect measure; Cohen's $d$ is naturally large for the
entropy-correlation comparisons ($d > 5$) because each ``observation''
is itself a mean over ${\sim}467$ test items, compressing variance; we
therefore rely on absolute deltas.
Results are verified across three independent data splits for robustness.

\section{Experimental Setup}
\label{sec:setup}

\paragraph{Model.} We use DeBERTa-v3-base \citep{he2023debertav3} (86M
parameters), pre-fine-tuned on MNLI, FEVER, and ANLI (full checkpoint
identifier in Appendix~\ref{app:implementation}).
Results with RoBERTa-large-mnli (355M parameters; Appendix~\ref{app:roberta})
replicate the qualitative separation with an even larger gap
($\Delta r = +0.249$ vs.\ $+0.154$).

\paragraph{Training.} We train for 5 epochs with effective batch size 16,
AdamW \citep{loshchilov2019decoupled}
($\text{lr} = 2 \times 10^{-5}$, weight decay 0.01),
maximum sequence length 128 tokens, and dynamic padding.
The best checkpoint is selected by validation loss (each mode uses its own
training objective). To ensure fair comparison, we apply temperature
scaling to \emph{all} models (not just the hard-label baseline),
providing a calibration-equalized comparison alongside raw results.

\section{Results}
\label{sec:results}

\subsection{Label Smoothing Cannot Substitute for Soft Labels}
\label{sec:gatekeeping}

\begin{figure}[t]
\centering
\includegraphics[width=\columnwidth]{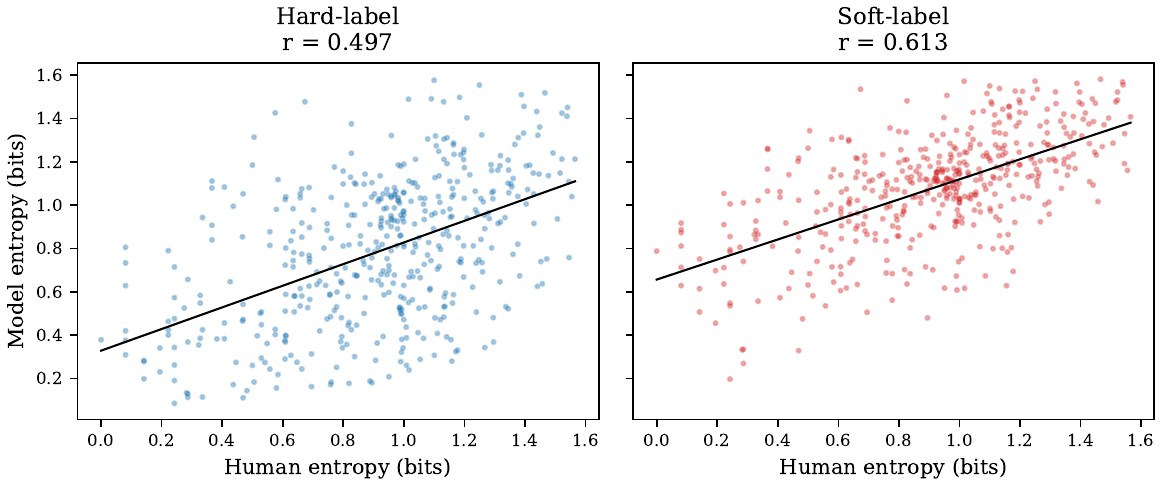}
\caption{Entropy correlation scatter plots for a single representative seed.
  Hard-label training (left) achieves $r = 0.497$; soft-label training (right)
  reaches $r = 0.613$, with tighter clustering along the diagonal.
  Five-seed means ($r = 0.487$ and $r = 0.643$; Table~\ref{tab:main})
  differ slightly because this seed is not exactly at the average.}
\label{fig:entropy_sep}
\end{figure}

Table~\ref{tab:main} presents the main comparison across seven training
configurations plus a temperature-scaled hard baseline, averaged over
five random seeds.
Figure~\ref{fig:entropy_sep} visualizes the central finding.

\begin{table*}[t]
\centering
\small
\begin{tabular}{lccc|cccc}
\toprule
& \multicolumn{3}{c|}{\textit{Majority-vote reference}} & \multicolumn{4}{c}{\textit{Distribution-aware (100-annotator reference)}} \\
\textbf{Config} & \textbf{Acc}$\uparrow$ & \textbf{ECE}$\downarrow$ & & \textbf{Brier-soft}$\downarrow$ & \textbf{Dist-ECE}$\downarrow$ & \textbf{KL}$\downarrow$ & \textbf{Ent.\ $r$}$\uparrow$ \\
\midrule
Hard-label         & .769\tiny{$\pm$.016} & \textbf{.036}\tiny{$\pm$.013} & & .118\tiny{$\pm$.006} & .090\tiny{$\pm$.005} & .226\tiny{$\pm$.015} & .487\tiny{$\pm$.018} \\
Hard + TempScale   & .769\tiny{$\pm$.016} & .038\tiny{$\pm$.012}$^\dagger$ & & .113\tiny{$\pm$.006} & .083\tiny{$\pm$.007} & .212\tiny{$\pm$.008} & .491\tiny{$\pm$.020} \\
\midrule
LS $\alpha{=}0.05$ & .765\tiny{$\pm$.019} & .047\tiny{$\pm$.010} & & .109\tiny{$\pm$.006} & .077\tiny{$\pm$.006} & .199\tiny{$\pm$.012} & .489\tiny{$\pm$.018} \\
LS $\alpha{=}0.1$  & .762\tiny{$\pm$.017} & .049\tiny{$\pm$.014} & & .103\tiny{$\pm$.006} & .064\tiny{$\pm$.008} & .182\tiny{$\pm$.010} & .487\tiny{$\pm$.018} \\
LS $\alpha{=}0.2$  & .764\tiny{$\pm$.020} & .077\tiny{$\pm$.031} & & .098\tiny{$\pm$.010} & .052\tiny{$\pm$.016} & .175\tiny{$\pm$.020} & .483\tiny{$\pm$.013} \\
LS $\alpha{=}0.3$  & .774\tiny{$\pm$.007} & .099\tiny{$\pm$.041} & & .095\tiny{$\pm$.005} & .050\tiny{$\pm$.010} & .177\tiny{$\pm$.009} & .460\tiny{$\pm$.027} \\
LS $\alpha{=}0.5$  & .775\tiny{$\pm$.006} & .201\tiny{$\pm$.034} & & .103\tiny{$\pm$.008} & .068\tiny{$\pm$.014} & .201\tiny{$\pm$.011} & .446\tiny{$\pm$.025} \\
\midrule
\textbf{Soft-label} & \textbf{.778}\tiny{$\pm$.006} & .098\tiny{$\pm$.010} & & \textbf{.079}\tiny{$\pm$.003} & \textbf{.031}\tiny{$\pm$.006} & \textbf{.136}\tiny{$\pm$.004} & \textbf{.643}\tiny{$\pm$.017} \\
\bottomrule
\end{tabular}
\caption{Main results (mean $\pm$ std over 5 seeds). Bold indicates best
  per column. The vertical rule separates the two evaluation regimes.
  \textit{Left}: majority-vote metrics, where ECE is best for hard-label
  because it penalizes models that express calibrated uncertainty on
  ambiguous items.
  \textit{Right}: distribution-aware metrics evaluated against the
  100-annotator human distribution, on which soft-label training
  dominates across all four measures.
  $^\dagger$Hard+TS ECE is slightly \emph{worse} than Hard (0.038 vs.\ 0.036)
  because temperature is optimized for NLL, not ECE; the found temperature
  ($T {\approx} 1.05$) marginally softens predictions, improving KL (0.212
  vs.\ 0.226) at the cost of a small ECE increase.
  For soft and LS models, temperature is fit by minimizing validation
  KL (Appendix~\ref{app:implementation}); a global temperature does not
  close the LS--soft gap (\S\ref{sec:analysis}).}
\label{tab:main}
\end{table*}

The primary result is a \emph{qualitative} separation on entropy correlation.
All five label-smoothing configurations cluster in a narrow band
($r \in [0.446, 0.489]$), while soft-label training reaches $r = 0.643$.
The absolute gap ($\Delta r = 0.154$--$0.197$) is consistent across all
five LS configurations (raw $p < 0.001$, all significant under
Holm-Bonferroni correction).
No tested $\alpha$ closes the gap.

For KL divergence, soft labels achieve 0.136 vs.\ the best label smoothing
at 0.175 ($\Delta = 0.039$; $p = 0.015$ uncorrected---the comparison
against LS $\alpha{=}0.3$, 0.136 vs.\ 0.177, is significant under Holm
correction, raw $p < 0.001$), a 22\% relative improvement. On
Brier-soft, soft labels achieve 0.079 vs.\ 0.095 ($\Delta = 0.016$,
$p = 0.003$ uncorrected).
Accuracy is slightly higher for soft labels (0.778 vs.\ 0.769), but the
difference is not statistically significant ($\Delta = 0.009$, $p = 0.13$).

\paragraph{Why ECE favors hard labels.} Standard ECE evaluates against
majority-vote labels. A model that outputs $P(\text{entailment}) = 0.62$
on a majority-entailment item is penalized as ``underconfident.''
The vertical rule in Table~\ref{tab:main} makes this separation explicit:
ECE uses a different evaluation reference than the distribution-aware
metrics. For models trained to match human distributions, only the
right-hand metrics are appropriate.

\paragraph{Why entropy correlation is the clearest signal.}
KL divergence and Brier-soft improve with label smoothing (LS $\alpha = 0.2$
achieves KL = 0.175, down from 0.226 for hard labels), because
\emph{any} softening of predictions reduces distance to the human
distribution. Entropy correlation, however, poses a more discriminating
question: does the model identify \emph{which items} are ambiguous?
Label smoothing provides no signal for this---every item's target has
identical entropy, so smoothing supplies no item-specific ranking
information, and empirically the rank ordering of model uncertainty
stays largely unchanged ($r$ near 0.45--0.49 at every intensity).
Only genuine soft labels provide the item-specific training
signal needed to learn disambiguation patterns.
The LS range ($r \in [0.446, 0.489]$) notably includes
values comparable to the hard-label baseline ($r = 0.487$), confirming
that uniform smoothing adds no measurable uncertainty-ranking signal.

\subsection{Stratified Analysis: The Cost of Item-Agnostic Smoothing}
\label{sec:stratified}

The introduction raised the hypothesis that label smoothing might
approximate soft labels on ambiguous items. Table~\ref{tab:stratified}
shows this holds only at high smoothing intensity, and at a steep cost
on clear items.

\begin{table}[t]
\centering
\small
\begin{tabular}{lccc}
\toprule
\textbf{Config} & \textbf{Low} & \textbf{Med} & \textbf{High} \\
& ($n{=}157$) & ($n{=}154$) & ($n{=}156$) \\
\midrule
Hard-label & .158 & .222 & .299 \\
Soft-label & \textbf{.121} & \textbf{.121} & \textbf{.167} \\
LS $\alpha{=}0.3$ & .177 & .177 & .177 \\
LS $\alpha{=}0.5$ & .265 & .196 & .143 \\
\bottomrule
\end{tabular}
\caption{KL divergence by human entropy tercile (mean over 5 seeds;
  per-cell std $\leq$ 0.034).
  Label smoothing at $\alpha = 0.3$ is nearly constant across terciles
  (KL $\approx$ 0.177 throughout), and soft labels match or beat it on
  every tercile (high-entropy 0.167 vs.\ 0.177, n.s., paired $p = 0.56$). Only at $\alpha = 0.5$ does uniform smoothing edge out
  soft labels on the highest-entropy tercile (0.143 vs.\ 0.167), at the cost
  of severe degradation on clear items (0.265 vs.\ 0.121). Soft labels track
  item difficulty and achieve the best aggregate performance. Bold marks the
  per-tercile best among Hard/Soft/LS~$\alpha{=}0.3$.}
\label{tab:stratified}
\end{table}

The mechanism is \emph{item-agnosticism}, the fundamental
limitation of uniform smoothing.
With $\alpha = 0.3$, 30\% of probability mass is redistributed
identically for every item, whether annotators agree or not,
producing near-identical error on clear and ambiguous items alike,
while soft labels track item difficulty (0.121 low-entropy, 0.167
high). Only at $\alpha = 0.5$ does the uniform target come close
enough to edge out soft labels on the highest-entropy tercile, while
error on clear items rises to over twice the soft-label level.

This item-agnosticism also explains why entropy correlation is
insensitive to $\alpha$: label smoothing shifts all items' target
uncertainty equally, adding no item-specific ranking signal, so
entropy correlation remains
near 0.45--0.49 regardless of smoothing intensity---consistent with
\citet{xia2025understanding}, who show via gradient analysis that
label smoothing degrades uncertainty rank ordering by regularizing
the max logit more for correct predictions.

\subsection{Annotation Efficiency: How Many Annotators?}
\label{sec:curve}

\begin{figure}[t]
\centering
\includegraphics[width=\columnwidth]{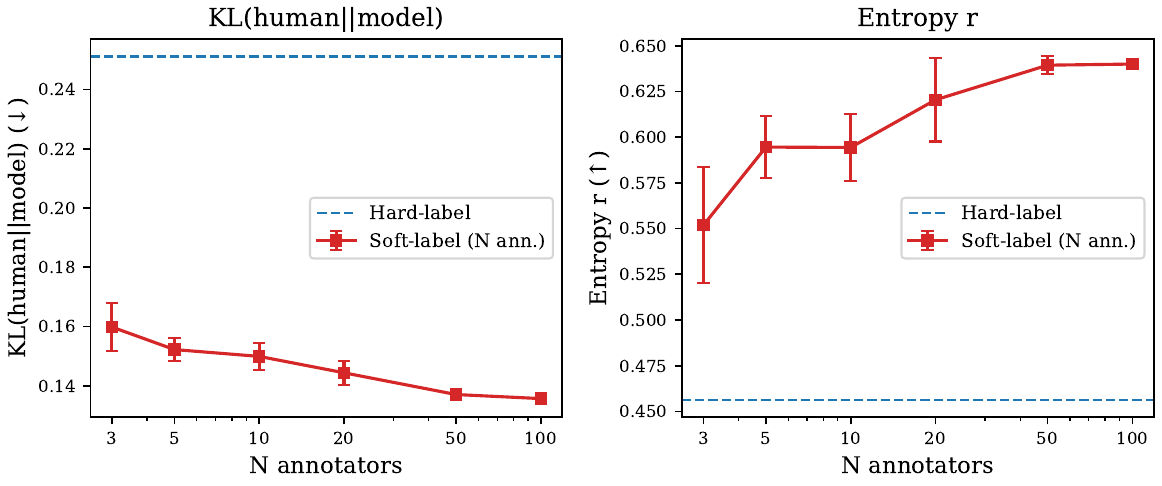}
\caption{Annotation efficiency curve showing the percentage of
  hard$\rightarrow$soft(100) improvement captured at each annotator count $N$.
  KL divergence (distributional match) saturates rapidly, while entropy
  correlation (uncertainty ranking) continues improving through $N \approx 50$.
  Error bars: $\pm 1$ std over 5 model seeds
  (each averaging 5 subsampling seeds).}
\label{fig:efficiency}
\end{figure}

Figure~\ref{fig:efficiency} and Table~\ref{tab:efficiency} present the
annotation efficiency curve, measuring the fraction of
hard$\rightarrow$soft(100) improvement captured at each annotator count $N$.

\begin{table}[t]
\centering
\small
\begin{tabular}{lcccc}
\toprule
$N$ & \textbf{KL}$\downarrow$ & \textbf{Ent.\ $r$}$\uparrow$ & \textbf{\% KL} & \textbf{\% Ent.\ $r$} \\
\midrule
Hard & .232\tiny{$\pm$.019} & .482\tiny{$\pm$.027} & 0\% & 0\% \\
3    & .154\tiny{$\pm$.006} & .569\tiny{$\pm$.014} & 81\tiny{$\pm$2}\% & 51\tiny{$\pm$5}\% \\
5    & .149\tiny{$\pm$.003} & .588\tiny{$\pm$.011} & 85\tiny{$\pm$1}\% & 62\tiny{$\pm$9}\% \\
10   & .144\tiny{$\pm$.004} & .608\tiny{$\pm$.009} & 91\tiny{$\pm$3}\% & 74\tiny{$\pm$6}\% \\
20   & .140\tiny{$\pm$.003} & .633\tiny{$\pm$.011} & 94\tiny{$\pm$3}\% & 89\tiny{$\pm$5}\% \\
50   & .137\tiny{$\pm$.002} & .648\tiny{$\pm$.006} & 98\tiny{$\pm$2}\% & 98\tiny{$\pm$3}\% \\
100  & .135\tiny{$\pm$.002} & .652\tiny{$\pm$.007} & 100\% & 100\% \\
\bottomrule
\end{tabular}
\caption{Efficiency curve results (5 model seeds $\times$ 5 subsampling seeds).
  Mean $\pm$ std across model seeds. The ``\%'' columns show the percentage of
  hard$\rightarrow$soft(100) improvement captured.
  KL saturates by $N \approx 10$ (91$\pm$3\%), while entropy $r$ lags behind
  at every $N < 20$ with wider variance.}
\label{tab:efficiency}
\end{table}

The central finding is \emph{differential saturation}: the two metrics
saturate at different rates.
A paired $t$-test at $N{=}3$---the point of maximum divergence---across
5 model seeds (each averaging over 5 subsampling seeds)
confirms that KL captures a significantly larger share of improvement
than entropy correlation ($\overline{\Delta} = 30.2$ pp, $t(4) = 17.7$, $p < 10^{-4}$,
one-sided). The gap remains significant at $N{=}10$
(16.7\,pp, $p = 0.0005$) and $N{=}20$ (5.1\,pp, $p = 0.013$),
narrowing to non-significance by $N{=}50$
(0.6\,pp, $p = 0.35$; full per-$N$ tests in
Appendix~\ref{app:spearman}).

\paragraph{KL divergence saturates rapidly.}
Three annotators capture 81\% of the full improvement; by $N = 10$,
91\% is captured. KL divergence is dominated by the
majority-class probability, which can be estimated reliably from
few annotators, explaining the early plateau.

\paragraph{Entropy correlation lags behind.}
Three annotators capture only 51\% of the improvement on entropy correlation,
compared to 81\% for KL---a 30 percentage point gap.
The gap narrows with more annotators but remains significant through $N{=}20$
(94\% vs.\ 89\%, $p = 0.013$). Individual model seeds show varying
trajectories: some exhibit an $N{=}5$--$10$ plateau, others improve
steadily (Appendix~\ref{app:seed43}), but the core pattern---entropy
correlation requiring more annotators than KL---is consistent across all five
model seeds, with convergence by $N \approx 50$.

\paragraph{Practical implications.}
For applications requiring distributional accuracy
(minimizing KL), 10 annotators per item may suffice (87--95\% of improvement
across five model seeds). For applications requiring uncertainty ranking---selective
prediction, active learning, or curriculum design---entropy correlation
lags behind KL at all $N < 20$ ($p < 0.001$ at $N{=}10$).
In our 3-class NLI setting, 20 annotators per item capture 89$\pm$5\% of
entropy correlation improvement at 20\% of the full annotation cost;
whether this threshold transfers to other label cardinalities and
tasks remains an open question.

\subsection{Cross-Domain Replication: DICES-990}
\label{sec:dices_main}

To test whether differential saturation is an NLI artifact, we run the
same efficiency-curve experiment on DICES-990 \citep{aroyo2024dices},
a content-safety dataset with 990 items, 2 classes (safe/unsafe), and
at least 58 annotators per item (we subsample up to $N{=}58$, the
minimum pool size), using DeBERTa-v3-base without safety pre-training
and 5 subsampling seeds.

\begin{table}[t]
\centering
\small
\begin{tabular}{lcccc}
\toprule
$N$ & \textbf{KL}$\downarrow$ & \textbf{Ent.\ $r$}$\uparrow$ & \textbf{\% KL} & \textbf{\% Ent.\ $r$} \\
\midrule
Hard & .323 & .440 & 0\% & 0\% \\
3    & .082\tiny{$\pm$.009} & .645\tiny{$\pm$.043} & 90\% & 62\% \\
5    & .073\tiny{$\pm$.004} & .669\tiny{$\pm$.041} & 93\% & 69\% \\
10   & .065\tiny{$\pm$.008} & .715\tiny{$\pm$.033} & 96\% & 83\% \\
20   & .061\tiny{$\pm$.003} & .736\tiny{$\pm$.027} & 97\% & 90\% \\
50   & .058\tiny{$\pm$.003} & .747\tiny{$\pm$.021} & 98\% & 93\% \\
58$^\dagger$ & .054 & .771 & 100\% & 100\% \\
\bottomrule
\end{tabular}
\caption{DICES-990 efficiency curve (DeBERTa-v3-base, 5 subsampling seeds).
  Differential saturation replicates: KL saturates faster (90\% at $N{=}3$)
  than entropy $r$ (62\% at $N{=}3$).
  Unlike ChaosNLI, no $N{=}5$--$10$ plateau is observed.
  $^\dagger$Single run: $N{=}58$ uses the full annotator pool, so
  subsampling seeds coincide.}
\label{tab:dices}
\end{table}

The same qualitative pattern holds (Table~\ref{tab:dices}): at
$N{=}3$, 90\% of the KL improvement is captured but only 62\% of
entropy correlation---a 28-point differential matching the ChaosNLI
pattern---and the gap persists through $N{=}20$ (97\% vs.\ 90\%).
We label this replication \emph{exploratory} because six design
differences (class count, annotator pool, task, item count,
pre-training, class balance) prevent clean attribution to task domain;
Appendix~\ref{app:dices} analyzes these confounds.

\subsection{Robustness Across Data Splits}
\label{sec:robustness}

To verify that our findings are not artifacts of a particular train/test split,
we repeat the comparison experiment across three independent data splits
(split seeds 42, 43, 44) with 2--5 model seeds each.

Across all three splits (5 seeds for split 42, 2 seeds each for splits
43 and 44; 9 seed-split combinations total), soft-label training wins
every comparison against the best label-smoothing configuration on KL
divergence, entropy correlation, and Brier-soft (9/9).
Accuracy is mixed (2/9), consistent with the non-significant accuracy
difference reported in \S\ref{sec:gatekeeping}. Dist-ECE favors soft
labels in 6/9 combinations (67\%).
The relative improvements on the distribution-aware metrics range from
37--53\% on KL and 22--47\% on entropy correlation across splits
(Appendix~\ref{app:splits}).

\section{Analysis}
\label{sec:analysis}

We now investigate the mechanisms behind differential saturation and validate the main findings through decomposition, held-out evaluation, and cross-architecture replication.

\paragraph{Seed dependence of the entropy correlation trajectory.}
The $N{=}5$--$10$ region shows the highest inter-seed variability
(\%ent.\ $r$ std: 9\,pp at $N{=}5$, 6\,pp at $N{=}10$), with some
seeds exhibiting a plateau and others improving steadily
(Appendix~\ref{app:seed43}). Minority-class MAE between subsamples
and the 100-annotator reference improves monotonically
($0.097 \to 0.068$ from $N{=}5$ to $N{=}10$), but whether the
model exploits this depends on initialization. By $N{=}20$, all
five seeds converge ($89{\pm}5$\%).

\paragraph{Why differential saturation is genuine.}
A potential concern is that the slower saturation of entropy correlation
is a statistical artifact---small-$N$ subsamples are noisier, and noise
may disproportionately affect correlation metrics. However, the differential
pattern is stable across all five subsampling seeds, all five model seeds
(42--46), both architectures (DeBERTa and RoBERTa), and is consistent
with the mathematical properties of the two metrics.

KL divergence is dominated by the \emph{largest} probability in the
distribution (the majority class), which can be estimated reliably from
few annotators. Entropy correlation depends on the \emph{full shape} of
the distribution, particularly the relative magnitudes of minority classes,
which require more annotators to resolve. The differential reflects the
differing information requirements of the two metrics: estimating an
item's majority-class mass from $N$ draws converges much faster than
estimating its entropy (subsampled-label entropy correlates with
100-annotator entropy at $r = 0.44/0.74/0.85$ for $N = 3/10/20$,
tracking the model curve).
Dirichlet smoothing
narrows the raw gap at $N{=}3$ (Appendix~\ref{app:dirichlet}),
confirming that estimation noise inflates its magnitude at very small
$N$, but the ordering persists at every $N$. Differential saturation replicates
on RoBERTa-large-mnli with a 32\,pp gap at $N{=}3$
(Table~\ref{tab:roberta-eff}), ruling out architecture-specific artifacts,
and on DICES-990 (2-class content safety; \S\ref{sec:dices_main})
with a 28\,pp gap, suggesting the pattern extends beyond NLI.

\paragraph{Temperature scaling as a diagnostic.}
A global temperature cannot close the LS--soft gap. With $T$ fit by
minimizing validation KL (Appendix~\ref{app:implementation}), soft-model
temperatures come out near-neutral ($T \in [1.16, 1.26]$ in a
verification re-run), and TS leaves soft-label results essentially
unchanged (KL 0.135$\to$0.127; entropy $r$ 0.652$\to$0.654)---the
distributions are already well-matched to human targets. The best
label-smoothing configuration after TS reaches only KL 0.161 and
entropy $r$ 0.494, far from soft-label performance: a single scalar
applied to every item can rescale confidence but does not create the
\emph{item-specific} uncertainty signal that entropy correlation
measures. Combined with the stratified analysis
(\S\ref{sec:stratified}), the evidence indicates that soft labels
learn item-specific distributional structure that global
temperature scaling does not reproduce.

\paragraph{Brier decomposition reveals the resolution gap.}
Table~\ref{tab:decomp} decomposes Brier-soft (Eq.~\ref{eq:murphy}).
Label smoothing reduces \emph{reliability} (0.033$\to$0.012) but
barely improves \emph{resolution} (0.160$\to$0.161); soft labels
improve both (REL=0.005, RES=0.171). Resolution measures exactly
what entropy correlation captures: sorting items by ambiguity.

\begin{table}[t]
\centering
\small
\begin{tabular}{lccc}
\toprule
\textbf{Config} & \textbf{REL}$\downarrow$ & \textbf{RES}$\uparrow$ & \textbf{UNC} \\
\midrule
Hard-label         & .033\tiny{$\pm$.003}  & .160\tiny{$\pm$.004}  & .247 \\
LS $\alpha{=}0.3$  & .012\tiny{$\pm$.004}  & .161\tiny{$\pm$.001}  & .247 \\
\textbf{Soft-label} & \textbf{.005}\tiny{$\pm$.002}  & \textbf{.171}\tiny{$\pm$.003}  & .247 \\
\bottomrule
\end{tabular}
\caption{Murphy decomposition of distributional Brier score
  (mean $\pm$ std, 5 seeds). REL = calibration error within bins
  (lower is better). RES = resolution across bins (higher is
  better). UNC = irreducible human disagreement variance (fixed).
  Soft-label training achieves both the lowest REL (best calibrated)
  and the highest RES (best at sorting items by ambiguity).
  Label smoothing reduces REL substantially (0.012 vs.\ 0.033) but
  barely improves RES (0.161 vs.\ 0.160), confirming that uniform
  smoothing calibrates on average without learning \emph{which} items
  are ambiguous.}
\label{tab:decomp}
\end{table}

\paragraph{Held-out annotator evaluation.}
To address evaluation circularity, we split annotators into 80
for training and 20 held-out for evaluation (5 seeds).
Soft labels still dominate on held-out annotators:
KL = 0.171 vs.\ 0.212 (best LS),
entropy $r$ = 0.577 vs.\ 0.417 (best LS; hard baseline 0.425).
The qualitative separation persists ($r \in [0.405, 0.417]$ for LS,
tested at $\alpha \in \{0.1, 0.3\}$, vs.\ $0.577$ for soft).

\paragraph{Cross-architecture replication.}
The qualitative separation replicates on RoBERTa-large-mnli
(Appendix~\ref{app:roberta}): LS entropy $r$ clusters at
$[0.351, 0.366]$ while soft labels reach $0.615$
($\Delta = +0.249$), an even larger gap than DeBERTa's $+0.154$.
DICES-990 (\S\ref{sec:dices_main}) adds a cross-domain replication,
with confounds analyzed in Appendix~\ref{app:dices}.

\paragraph{Pre-training ablation.}
DeBERTa-v3-base was pre-fine-tuned on MNLI/FEVER/ANLI with hard labels.
To rule out this confound, we repeat the comparison using
DeBERTa-v3-base without NLI pre-training
(Appendix~\ref{app:pretrain}).
The entropy correlation gap \emph{widens} ($+0.154 \to +0.175$),
confirming the advantage is not an initialization artifact.

\paragraph{Dirichlet smoothing.}
Adding pseudo-counts to label counts mitigates zero-count issues
at small $N$ (Appendix~\ref{app:dirichlet}).
At $N{=}10$ with $\alpha{=}1/K$, smoothed KL nearly matches
$N{=}100$ (0.137 vs.\ 0.136), closing 99\% of the gap; smoothing
significantly improves KL over raw counts at $N{=}10$ ($p = 0.004$)
while leaving entropy correlation statistically unchanged
($p = 0.12$, ns; 87\% of the gap closed).
Differential saturation persists under smoothing.

\section{Conclusion}
\label{sec:conclusion}

Our results on ChaosNLI show that different evaluation objectives
have different annotation requirements:
distributional match (KL) requires only
$N \approx 10$ annotators (87--95\% of improvement),
while uncertainty ranking (entropy correlation) requires
$N \approx 20$--$50$---a pattern we term
\emph{metric-dependent saturation}.
This rests on a prerequisite: soft labels carry item-specific signal
that label smoothing cannot replicate ($\Delta r = 0.154$, $p < 0.001$),
a gap that replicates across architectures and a non-NLI-pretrained
initialization; differential saturation additionally appears in an
exploratory content-safety evaluation.
The value of real annotations lies in their
\emph{item-specificity}---and the amount needed depends on
what you intend to measure.

\paragraph{Implications.}
First, annotation budgets benefit from specifying the
downstream metric: in our NLI setting, 10 annotators suffice for
distributional match, while entropy correlation converges only in the
$N{=}20$--$50$ range depending on initialization
(Appendix~\ref{app:seed43}).
Second, distribution-aware metrics such as entropy correlation and
Brier-soft provide information that accuracy and ECE obscure when the
evaluation target is a distribution rather than a point label.
Third, label-smoothing baselines offer a useful zero-cost control
for studies of soft-label training.
Fourth, Dirichlet smoothing partially substitutes
for additional annotations---at $N{=}10$ with $\alpha{=}1/K$,
smoothed KL closes 99\% of the hard-to-soft gap ($p = 0.004$ vs.\ raw
counts) while entropy correlation closes only 87\% (ns, $p = 0.12$),
so differential saturation persists even with better aggregation.
Fifth, annotator models that assume a single true label
(e.g., Dawid-Skene; Appendix~\ref{app:dawid_skene}) collapse
distributional signal; raw counts retain \emph{how people disagree}.

\paragraph{Future work.}
For \emph{ordinal} labels, where disagreement partly reflects scale
usage, distance-aware smoothing may close more of the
distribution-match gap, though it still cannot identify ambiguous
items; graded-label datasets and decoder-only LLMs are the natural
next tests.

\section*{Limitations}

\paragraph{Primarily NLI.}
Our main experiments use ChaosNLI (MNLI + SNLI, 3,113 items, 3-class NLI).
The exploratory replication on DICES-990 (\S\ref{sec:dices_main}) suggests
differential saturation generalizes to content safety, but this comparison
is confounded by multiple design differences (Appendix~\ref{app:dices}).
The specific saturation thresholds ($N \approx 10$ for KL,
$N \approx 20$--$50$
for entropy correlation) may not transfer to other tasks with different
label sets or annotator pools.

\paragraph{Two architectures, two initialization conditions.}
We use DeBERTa-v3-base as our primary model. Results with RoBERTa-large-mnli
(Appendix~\ref{app:roberta}) confirm the qualitative separation replicates
with an even larger gap. A pre-training ablation using DeBERTa-v3-base without NLI
pre-training (Appendix~\ref{app:pretrain}) shows the gap widens
rather than shrinks, ruling out initialization confounds.
However, both architectures are encoder-only transformers, the
standard setting in the learning-from-disagreement literature we build
on. Whether the pattern holds for decoder-only and instruction-tuned
LLMs fine-tuned on label distributions, and whether the specific
saturation thresholds differ without NLI pre-training, are explicit
directions for future work.

\paragraph{When hard labels may suffice.}
As human entropy approaches zero (unanimous agreement), soft labels
converge to hard labels. For datasets with predominantly unambiguous
items, hard-label training may be equally effective, and the cost of
collecting multiple annotations per item would be wasted.

\paragraph{Small training set.}
The training set (2,179 items) is small by modern standards. The strong
pre-training of DeBERTa-v3-base (on MNLI/FEVER/ANLI) compensates, but
effects may differ with larger training sets.

\paragraph{Early stopping and evaluation alignment.}
Each training mode uses its own validation loss for early stopping,
following standard practice: each model gets its best checkpoint
under the objective it was trained on. Since the soft-label loss
(KL divergence) aligns more closely with distributional evaluation
metrics than cross-entropy does, we apply temperature scaling to all
models as a calibration-equalized fairness control (\S\ref{sec:setup}).
Soft-label models are trained on and evaluated against the same
100-annotator pool, but on disjoint item sets (train/test split).
The evaluation measures generalization to \emph{unseen items} from
the same annotator population---standard practice in ML---but
results may differ with a different annotator pool.

\paragraph{Oracle validation distributions.}
Our efficiency curves subsample \emph{training} labels only; validation
distributions for early stopping always use all 100 annotators. This
isolates training-signal saturation but means a fully $N$-budgeted
pipeline---including validation annotation---may capture less of the
improvement than reported.

\paragraph{Items versus annotators.}
The design holds the item set fixed to isolate the annotator axis: the
curves answer ``given these items, how many annotations per item,''
not ``how should a fixed budget be split between more items and more
annotations per item.'' Prior work has compared breadth-vs-depth
allocations directly \citep{zhang2021capturing}; mapping the joint
budget frontier is an open question.

\paragraph{Statistical precision.}
We use 5 model seeds for both the main comparison and the efficiency
curve (5 model seeds $\times$ 5 subsampling seeds = 25 observations per $N$).
The large effect sizes ($\Delta\text{KL} = 0.090$,
$\Delta r = 0.156$; gatekeeping comparison, soft vs.\ hard) make
directional findings robust, but exact saturation
thresholds carry uncertainty.

\paragraph{Hardware determinism.}
All experiments run on Apple M3 (MPS backend). MPS does not guarantee
bitwise reproducibility, but empirical testing shows that repeated
training with the same seed produces effectively identical results
(metric range $< 10^{-6}$ across five replicas on all eight metrics). Differences between
experiments reflect different model seeds, not hardware noise; we
mitigate seed variance through multi-seed averaging (5 model seeds,
5 subsampling seeds).

\section*{Ethics Statement}
This work uses the publicly available ChaosNLI dataset \citep{nie2020what},
released under CC-BY-NC 4.0, and the publicly available DICES dataset
\citep{aroyo2024dices}, released by Google Research under CC-BY 4.0;
DICES contains conversations rated for safety, some involving harmful
content, and we use only the aggregate rating distributions.
All experiments use open-source pre-trained models.
No new human annotations were collected.
Writing assistance was provided by a generative AI tool (Claude, Anthropic).

\bibliography{references}

\appendix

\section{Split-Seed Robustness Details}
\label{app:splits}

Table~\ref{tab:splits} reports the per-split hard$\to$soft
improvements underlying \S\ref{sec:robustness}: the gains are
consistent in direction and magnitude across all three data splits.

\begin{table}[h]
\centering
\small
\setlength{\tabcolsep}{5pt}
\begin{tabular}{lccc}
\toprule
\textbf{Split} & \textbf{KL} & \textbf{Ent.\ $r$} & \textbf{Brier-soft} \\
& (hard$\to$soft) & (hard$\to$soft) & (hard$\to$soft) \\
\midrule
42 (5 seeds) & .226$\to$.136 & .487$\to$.643 & .118$\to$.079 \\
& ($-$40\%) & (+32\%) & ($-$33\%) \\
43 (2 seeds) & .263$\to$.166 & .473$\to$.579 & .137$\to$.098 \\
& ($-$37\%) & (+22\%) & ($-$29\%) \\
44 (2 seeds) & .313$\to$.148 & .416$\to$.612 & .159$\to$.086 \\
& ($-$53\%) & (+47\%) & ($-$46\%) \\
\bottomrule
\end{tabular}
\caption{Soft vs.\ hard-label performance across three data splits.
  Improvements are consistent across splits, with relative KL reduction
  ranging from 37--53\% and entropy correlation improvement from 22--47\%.}
\label{tab:splits}
\end{table}

\section{Implementation Details}
\label{app:implementation}

\paragraph{Model checkpoint.} DeBERTa-v3-base is initialized from
\texttt{MoritzLaurer/\allowbreak DeBERTa-v3-base-mnli-fever-anli};
RoBERTa-large from \texttt{roberta-large-mnli}.

\paragraph{Training details.} We use the HuggingFace Transformers library
\citep{wolf2020transformers} with the Trainer API. Training uses AdamW
\citep{loshchilov2019decoupled} with $\beta_1 = 0.9$, $\beta_2 = 0.999$,
$\epsilon = 10^{-8}$, learning rate $2 \times 10^{-5}$, linear warmup
over the first 10\% of steps, and weight decay 0.01.

\paragraph{Soft-label loss.} For soft-label and label-smoothing
configurations, we use $\text{KL}(\mathbf{t} \| \text{softmax}(\mathbf{z}))$
where $\mathbf{t}$ is the target distribution and $\mathbf{z}$ are the
model logits. For hard-label training, KL with one-hot targets reduces to
standard cross-entropy.

\paragraph{Temperature scaling.} For the Hard+TempScale baseline, we train
a hard-label model and then optimize a single scalar temperature $T$ on
the validation set by minimizing NLL. The temperature is applied at test
time as $\mathbf{p} = \text{softmax}(\mathbf{z}/T)$.

\paragraph{Early stopping.} Each training mode uses its own validation
loss for best-checkpoint selection. To control for potential differences
in calibration arising from mode-specific stopping, we apply temperature
scaling to all models. For hard-label models, $T$ is optimized by
minimizing NLL against majority-vote labels; for soft-label and
label-smoothing models, $T$ is optimized by minimizing KL divergence
against the soft validation labels, preserving alignment with each
model's training objective.
The temperature-scaled comparison provides a calibration-equalized
baseline; see main text for both raw and temperature-scaled results.

\paragraph{Efficiency curve.} For the annotation efficiency experiment,
we run five model seeds (42--46), each with five subsampling seeds
(100--104), yielding 25 observations per $N$ (plus 5 hard baselines).
This captures both model initialization variance and annotator
subsampling variance. Crucially, only
training labels are subsampled---the validation set always uses the
full 100-annotator distributions for early stopping, ensuring
stopping quality is constant across $N$.
Hard-label baselines differ slightly between the gatekeeping
(Table~\ref{tab:main}) and efficiency curve
(Table~\ref{tab:efficiency}) experiments because
\texttt{dataloader\_num\_workers=2} introduces session-level
non-determinism in batch ordering; this is absorbed by
multi-seed averaging.

\paragraph{Data preprocessing.} Premises and hypotheses are concatenated
with the tokenizer's separator token. Dynamic padding pads each batch
to the length of its longest sequence rather than a fixed maximum length.

\section{RoBERTa-large-mnli Results}
\label{app:roberta}

To verify that our findings generalize beyond a single architecture, we
repeat the gatekeeping comparison (\S\ref{sec:gatekeeping}) using
RoBERTa-large-mnli (355M parameters) with 5 model seeds. Label remapping
between the model's internal label ordering and the canonical ChaosNLI
ordering is handled automatically.

\begin{table}[h]
\centering
\small
\setlength{\tabcolsep}{4pt}
\begin{tabular}{lcccc}
\toprule
\textbf{Config} & \textbf{Acc}$\uparrow$ & \textbf{Brier-soft}$\downarrow$ & \textbf{KL}$\downarrow$ & \textbf{Ent.\ $r$}$\uparrow$ \\
\midrule
Hard-label         & .747\tiny{$\pm$.012} & .154\tiny{$\pm$.012} & .356\tiny{$\pm$.036} & .353\tiny{$\pm$.017} \\
\midrule
LS $\alpha{=}0.05$ & .748\tiny{$\pm$.011} & .140\tiny{$\pm$.010} & .283\tiny{$\pm$.027} & .361\tiny{$\pm$.016} \\
LS $\alpha{=}0.1$  & .747\tiny{$\pm$.008} & .127\tiny{$\pm$.009} & .238\tiny{$\pm$.023} & .366\tiny{$\pm$.021} \\
LS $\alpha{=}0.2$  & .749\tiny{$\pm$.007} & .112\tiny{$\pm$.005} & .200\tiny{$\pm$.011} & .364\tiny{$\pm$.035} \\
LS $\alpha{=}0.3$  & .750\tiny{$\pm$.009} & .106\tiny{$\pm$.004} & .192\tiny{$\pm$.006} & .364\tiny{$\pm$.034} \\
LS $\alpha{=}0.5$  & .752\tiny{$\pm$.014} & .120\tiny{$\pm$.004} & .225\tiny{$\pm$.005} & .351\tiny{$\pm$.017} \\
\midrule
\textbf{Soft-label} & \textbf{.784}\tiny{$\pm$.008} & \textbf{.079}\tiny{$\pm$.002} & \textbf{.145}\tiny{$\pm$.002} & \textbf{.615}\tiny{$\pm$.022} \\
\bottomrule
\end{tabular}
\caption{RoBERTa-large-mnli results (mean $\pm$ std, 5 seeds).
  The qualitative separation replicates: LS clusters at $r \in [0.351, 0.366]$,
  while soft labels reach $r = 0.615$ ($\Delta = +0.249$).
  The gap is \emph{larger} than for DeBERTa ($\Delta = +0.154$),
  likely because RoBERTa's weaker NLI baseline ($r = 0.353$ vs.\ $0.487$)
  leaves more room for soft-label improvement.}
\label{tab:roberta}
\end{table}

The qualitative separation between soft labels and label smoothing
replicates cleanly (Table~\ref{tab:roberta}). All five LS configurations
cluster in $r \in [0.351, 0.366]$---an even tighter band than DeBERTa's
$[0.446, 0.489]$---while soft labels reach $r = 0.615$.
The absolute gap ($\Delta r = 0.249$) is larger than DeBERTa's ($0.154$),
consistent with the hypothesis that a weaker hard-label baseline provides
more room for distributional signal to improve uncertainty ranking.

Soft-label training also achieves the highest accuracy (0.784 vs.\ 0.752
for the best LS), demonstrating that distributional training does not
sacrifice classification performance. On KL divergence, the improvement
is even more pronounced: soft labels achieve 0.145 vs.\ the best LS
at 0.192 ($\Delta = 0.048$, 25\% relative improvement).

\paragraph{Efficiency curve replication.}
We also replicate the efficiency curve experiment (\S\ref{sec:efficiency})
on RoBERTa-large-mnli with 5 subsampling seeds (Table~\ref{tab:roberta-eff}).

\begin{table}[h]
\centering
\small
\begin{tabular}{lcccc}
\toprule
$N$ & \textbf{KL}$\downarrow$ & \textbf{Ent.\ $r$}$\uparrow$ & \textbf{\% KL} & \textbf{\% Ent.\ $r$} \\
\midrule
Hard$^\dagger$ & .366 & .356 & 0\% & 0\% \\
3    & .174\tiny{$\pm$.015} & .500\tiny{$\pm$.030} & 86\% & 54\% \\
5    & .161\tiny{$\pm$.010} & .526\tiny{$\pm$.038} & 92\% & 63\% \\
10   & .158\tiny{$\pm$.004} & .520\tiny{$\pm$.019} & 93\% & 61\% \\
20   & .152\tiny{$\pm$.005} & .566\tiny{$\pm$.010} & 96\% & 78\% \\
50   & .147\tiny{$\pm$.004} & .620\tiny{$\pm$.007} & 98\% & 98\% \\
100$^\dagger$ & .143 & .625 & 100\% & 100\% \\
\bottomrule
\end{tabular}
\caption{RoBERTa-large-mnli efficiency curve (5 subsampling seeds, model seed 42).
  Differential saturation replicates: at $N{=}3$, KL captures 86\%
  but entropy $r$ only 54\% (gap of 32\,pp vs.\ DeBERTa's 30\,pp).
  With this model seed (42), an $N{=}5$--$10$ entropy $r$ stall also
  appears ($r = 0.526 \to 0.520$); see Appendix~\ref{app:seed43} for
  seed dependence of this trajectory.
  $^\dagger$Single deterministic run (no std).}
\label{tab:roberta-eff}
\end{table}

Differential saturation replicates clearly: at $N{=}3$ the gap between
\% KL (86\%) and \% entropy $r$ (54\%) is 32 percentage points---comparable
to DeBERTa's 30\,pp gap (5-seed average), consistent with RoBERTa's weaker
hard-label baseline providing a wider range of improvement.
With this model seed (42), entropy $r$ also shows an $N{=}5$--$10$ stall
($r = 0.526 \to 0.520$), similar to DeBERTa seed 42.
However, several DeBERTa seeds do \emph{not} show this stall
(Appendix~\ref{app:seed43}), confirming it is seed-dependent rather
than a fundamental property of the $N{=}5$--$10$ range.

\section{Efficiency Curve: Per-Seed Details}
\label{app:seed43}

The main Table~\ref{tab:efficiency} reports means across five model seeds
(42--46). Here we show seed 43 in detail to illustrate seed-dependent
trajectory differences (all other parameters identical: 5 subsampling seeds,
split seed 42).

\begin{table}[h]
\centering
\small
\begin{tabular}{lcccc}
\toprule
$N$ & \textbf{KL}$\downarrow$ & \textbf{Ent.\ $r$}$\uparrow$ & \textbf{\% KL} & \textbf{\% Ent.\ $r$} \\
\midrule
Hard$^\dagger$ & .219 & .486 & 0\% & 0\% \\
3    & .146\tiny{$\pm$.003} & .579\tiny{$\pm$.009} & 85\% & 55\% \\
5    & .146\tiny{$\pm$.003} & .594\tiny{$\pm$.027} & 85\% & 64\% \\
10   & .139\tiny{$\pm$.004} & .619\tiny{$\pm$.026} & 93\% & 79\% \\
20   & .136\tiny{$\pm$.002} & .648\tiny{$\pm$.005} & 96\% & 96\% \\
50   & .133\tiny{$\pm$.002} & .655\tiny{$\pm$.005} & 100\% & 100\% \\
100$^\dagger$ & .133 & .655 & 100\% & 100\% \\
\bottomrule
\end{tabular}
\caption{DeBERTa efficiency curve with model seed 43 (5 subsampling seeds).
  Differential saturation replicates: at $N{=}3$, KL captures 85\%
  but entropy $r$ only 55\% (gap of 30\,pp).
  Unlike seed 42, entropy $r$ improves steadily from $N{=}5$ to $N{=}10$
  ($r = 0.594 \to 0.619$), with no plateau.
  $^\dagger$Single deterministic run (no std).}
\label{tab:seed43}
\end{table}

\paragraph{Differential saturation replicates across all seeds.}
The core finding is robust: KL saturates faster than entropy correlation
at every $N$ across all five model seeds. For seed 43, the gap at $N{=}3$
is 30\,pp (85\% KL vs.\ 55\% entropy $r$), consistent with the 5-seed
average of 30\,pp ($p < 10^{-4}$; Table~\ref{tab:pern_tests}).
Note that saturation percentages are relative to each seed's own
hard$\to$N{=}100 improvement, so the within-seed KL-vs-entropy
comparison is the meaningful contrast.

\paragraph{The $N{=}5$--$10$ trajectory is seed-dependent.}
Some seeds (e.g., 42) show a plateau at $N{=}5$--$10$ while others
(e.g., 43) show steady improvement ($r = 0.594 \to 0.619$). Across all
five seeds, the inter-seed std of \%ent.\ $r$ is highest at $N{=}5$
(9\,pp) and $N{=}10$ (6\,pp), confirming that the $N{=}5$--$10$ region
is where model stochasticity has the greatest effect on entropy
correlation---an important consideration for practitioners choosing
annotation budgets.

\paragraph{Confidence in saturation thresholds.}
The $N$ at which \%KL first exceeds 90\% is $N{=}10$ for 3 of 5 seeds
and $N{=}20$ for the remaining 2 (per-seed range at $N{=}10$:
87--95\%). The $N$ at which \%ent.\,$r$ first exceeds 90\% is $N{=}20$
for 2 seeds and $N{=}50$ for 3 seeds (per-seed range at $N{=}10$:
65--80\%). This makes the uncertainty in the threshold explicit:
``$N \approx 10$ for KL'' is a reasonable summary, while
entropy correlation's threshold falls in the $N{=}20$--$50$ range
depending on model initialization.

\section{Efficiency Curve: All Metrics}
\label{app:efficiency_full}

\begin{figure}[h]
\centering
\includegraphics[width=\columnwidth]{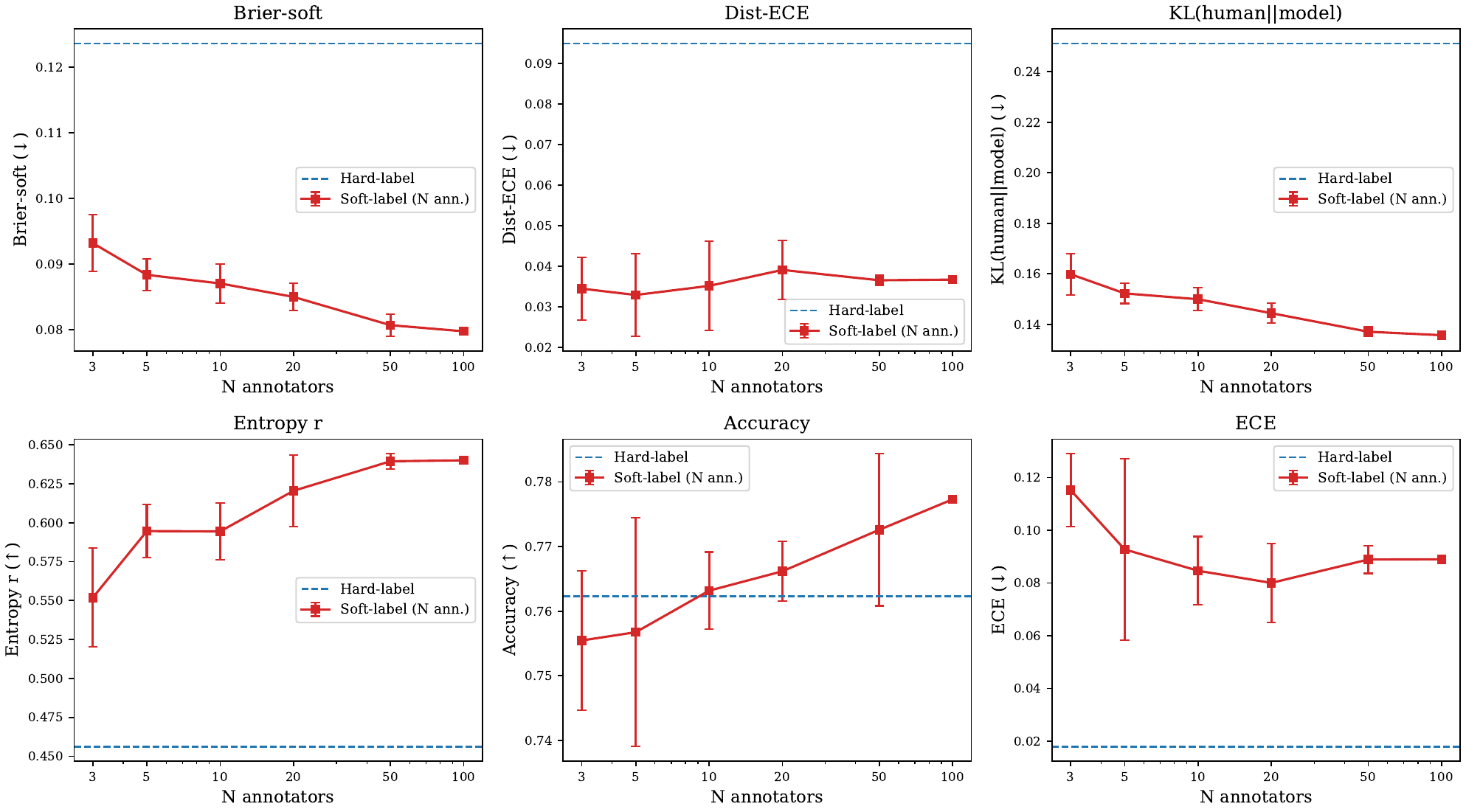}
\caption{Annotation efficiency curve for all six metrics
  (5 model seeds $\times$ 5 subsampling seeds). This is the
  full version of Figure~\ref{fig:efficiency}, which shows only
  KL and entropy $r$.}
\label{fig:efficiency_full}
\end{figure}

Figure~\ref{fig:efficiency_full} shows the annotation efficiency curve
for all six evaluation metrics. The main text (Figure~\ref{fig:efficiency})
displays only KL divergence and entropy correlation, the two metrics
central to the differential saturation finding. The remaining four
metrics (Brier-soft, Dist-ECE, accuracy, ECE) are included here for
completeness.

\section{DICES-990: Confounds and Details}
\label{app:dices}

The DICES-990 replication (\S\ref{sec:dices_main}, Table~\ref{tab:dices})
uses 10 training epochs; all other hyperparameters follow
Appendix~\ref{app:implementation}.

\paragraph{Important caveats.}
The cross-domain comparison is \emph{exploratory} and cannot cleanly
isolate the effect of task domain. DICES-990 differs from ChaosNLI in six ways:
(1)~2 vs.\ 3 classes, (2)~annotator pool ($\geq$58 vs.\ 100 per item),
(3)~990 vs.\ 3,113 items, (4)~content safety vs.\ NLI,
(5)~no task-specific pre-training, and (6)~strong class imbalance
(83.5\% unsafe by majority vote). Any of these could affect saturation dynamics.
In particular, the faster KL saturation on DICES (90\% at $N{=}3$
vs.\ 81\% on ChaosNLI) may reflect the 2-class setting, where the
majority-class probability is even easier to estimate with few annotators.

The absence of a $N{=}5$--$10$ plateau on DICES may similarly reflect
the simpler label space: with only two classes, there is no ``minority
class resolution'' problem---any annotator disagreement directly reveals
the safe/unsafe split.

We do not run the full label-smoothing comparison on DICES,
so we cannot confirm whether the LS--soft separation that motivates
the NLI findings also holds in this domain. A complete gatekeeping
experiment on DICES would strengthen the cross-domain claim.

\section{Dirichlet Smoothing Details}
\label{app:dirichlet}

To test whether Bayesian aggregation can substitute for collecting
additional annotations, we apply Dirichlet smoothing to subsampled
label counts before converting to training distributions.
Given observed counts $\mathbf{c} = (c_1, \ldots, c_K)$ from $N$
annotators, the smoothed distribution is the posterior mean of a
symmetric Dirichlet-multinomial:
\begin{equation}
\hat{p}_k = \frac{c_k + \alpha}{\sum_{j} c_j + K\alpha}\,.
\end{equation}
We test three principled $\alpha$ values: $1/K$ (uniform prior,
minimally informative), $0.5$ (Jeffreys prior), and $1.0$ (Laplace
prior). Training uses the same fixed model seed (42) and the same
five subsampling seeds (100--104) as the main efficiency curve,
enabling direct comparison.

\begin{table}[h]
\centering
\small
\begin{tabular}{llccc}
\toprule
\textbf{$N$} & \textbf{$\alpha$} & \textbf{KL}$\downarrow$ & \textbf{Ent.\ $r$}$\uparrow$ & \textbf{Acc} \\
\midrule
3 & raw (no smooth.) & .160\tiny{$\pm$.008} & .552\tiny{$\pm$.032} & .755 \\
3 & $1/K$ & .159\tiny{$\pm$.004} & .575\tiny{$\pm$.018} & .752 \\
3 & $0.5$ & .173\tiny{$\pm$.003} & .543\tiny{$\pm$.018} & .747 \\
3 & $1.0$ & .209\tiny{$\pm$.004} & .525\tiny{$\pm$.015} & .742 \\
\midrule
5 & raw (no smooth.) & .152\tiny{$\pm$.004} & .595\tiny{$\pm$.017} & .757 \\
5 & $1/K$ & .149\tiny{$\pm$.005} & .600\tiny{$\pm$.015} & .754 \\
5 & $0.5$ & .154\tiny{$\pm$.005} & .594\tiny{$\pm$.014} & .754 \\
5 & $1.0$ & .176\tiny{$\pm$.003} & .583\tiny{$\pm$.014} & .757 \\
\midrule
10 & raw (no smooth.) & .150\tiny{$\pm$.005} & .594\tiny{$\pm$.018} & .763 \\
10 & $1/K$ & .137\tiny{$\pm$.004} & .617\tiny{$\pm$.008} & .773 \\
10 & $0.5$ & .138\tiny{$\pm$.004} & .611\tiny{$\pm$.008} & .776 \\
10 & $1.0$ & .145\tiny{$\pm$.003} & .608\tiny{$\pm$.007} & .770 \\
\midrule
100 & raw (reference) & .136 & .640 & .777 \\
\bottomrule
\end{tabular}
\caption{Dirichlet smoothing at small annotator counts.
  The lightest prior ($\alpha{=}1/K$) consistently performs best.
  At $N{=}10$, smoothed KL nearly matches the $N{=}100$ reference
  (0.137 vs.\ 0.136), but entropy correlation still lags
  (0.617 vs.\ 0.640). Heavier priors hurt at $N{=}3$
  (ent.\ $r$ drops below the raw baseline).
  Differential saturation persists across all smoothing levels.
  ``Raw'' rows use the same seeds as the main efficiency curve
  (Table~\ref{tab:efficiency}).}
\label{tab:dirichlet}
\end{table}

Key observations: (1)~$\alpha{=}1/K$ dominates on KL at every $N$;
heavier priors ($\alpha{=}1.0$) over-smooth and hurt distributional
match, especially at small $N$. (2)~At $N{=}3$, only $\alpha{=}1/K$
shows a directional improvement over raw counts on entropy correlation
($0.575$ vs.\ $0.552$; not significant, $p = 0.13$);
$\alpha \geq 0.5$ actually hurts, as over-smoothing washes out the
sparse signal. (3)~At $N{=}10$, all three priors help and converge---smoothing
becomes less sensitive when the empirical distribution is already
reliable. The KL improvement at $N{=}10$ is significant
(paired $t(4) = -5.8$, $p = 0.004$); the mechanism is zero-count
mitigation for minority classes, which KL penalizes harshly.
(4)~Even with optimal smoothing ($\alpha{=}1/K$), entropy
correlation at $N{=}10$ closes 87\% of the hard-to-soft gap
while KL closes 99\%. The differential narrows at $N{=}3$
(from 27\,pp to 15\,pp), suggesting estimation noise contributes
to the raw gap at very small $N$, but differential saturation
persists in sign at all $N$.

Note that Dirichlet smoothing changes both the signal quality
(distributions closer to the 100-annotator reference) and the
loss landscape (non-zero gradients for all classes). We cannot
fully disentangle these effects; the improvement may partly
reflect regularization benefits beyond pure aggregation quality.

\section{Spearman Correlation and Per-$N$ Statistical Tests}
\label{app:spearman}

\paragraph{Spearman $\rho$ vs.\ Pearson $r$.}
All entropy correlations in the main text use Pearson $r$.
Table~\ref{tab:spearman} reports Spearman $\rho$ for the main
comparison and efficiency curve; the rank-order conclusions
are identical.

\begin{table}[h]
\centering
\small
\begin{tabular}{lcc}
\toprule
\textbf{Config} & \textbf{Pearson $r$} & \textbf{Spearman $\rho$} \\
\midrule
\multicolumn{3}{l}{\textit{Gatekeeping (DeBERTa, 5 seeds)}} \\
Hard-label       & .487\tiny{$\pm$.018} & .466\tiny{$\pm$.018} \\
LS $\alpha{=}0.1$  & .487\tiny{$\pm$.018} & .466\tiny{$\pm$.018} \\
LS $\alpha{=}0.3$  & .460\tiny{$\pm$.027} & .447\tiny{$\pm$.024} \\
\textbf{Soft-label} & \textbf{.643}\tiny{$\pm$.017} & \textbf{.611}\tiny{$\pm$.009} \\
\midrule
\multicolumn{3}{l}{\textit{Gatekeeping (RoBERTa, 5 seeds)}} \\
Hard-label       & .353\tiny{$\pm$.017} & .341\tiny{$\pm$.013} \\
\textbf{Soft-label} & \textbf{.615}\tiny{$\pm$.022} & \textbf{.574}\tiny{$\pm$.016} \\
\midrule
\multicolumn{3}{l}{\textit{Efficiency curve (DeBERTa, seed 42)}} \\
$N{=}3$          & .552\tiny{$\pm$.032} & .523\tiny{$\pm$.020} \\
$N{=}10$         & .594\tiny{$\pm$.018} & .565\tiny{$\pm$.014} \\
$N{=}50$         & .639\tiny{$\pm$.005} & .596\tiny{$\pm$.005} \\
\bottomrule
\end{tabular}
\caption{Pearson $r$ vs.\ Spearman $\rho$ for entropy correlation.
  Spearman values are consistently lower in magnitude but the
  qualitative separation between soft and LS/hard is identical.}
\label{tab:spearman}
\end{table}

\paragraph{Per-$N$ paired $t$-tests for differential saturation.}
Table~\ref{tab:pern_tests} tests whether \%KL $>$ \%Ent.\,$r$
at each annotator count. Each of the 5 model seeds contributes one
observation (mean across 5 subsampling seeds), yielding a conservative
df$=$4 test.

\begin{table}[h]
\centering
\small
\begin{tabular}{rrrrl}
\toprule
$N$ & \%KL & \%Ent.\,$r$ & $p$ & \\
\midrule
3  & 81$\pm$2  & 51$\pm$5  & $<$.0001 & *** \\
5  & 85$\pm$1  & 62$\pm$9   & .002  & ** \\
10 & 91$\pm$3  & 74$\pm$6  & .0005 & *** \\
20 & 94$\pm$3  & 89$\pm$5  & .013  & * \\
50 & 98$\pm$2  & 98$\pm$3  & .35  & ns \\
\bottomrule
\end{tabular}
\caption{Per-$N$ paired $t$-tests (df$=$4, one-sided, 5 model seeds).
  The gap is significant at $N{=}3$, $5$, $10$, and $20$, converging
  by $N{=}50$.}
\label{tab:pern_tests}
\end{table}

\section{Annotator Modeling and the Single-Truth Assumption}
\label{app:dawid_skene}

Dawid-Skene (DS; \citealt{dawid1979maximum}) is the canonical
annotator model: it posits a single latent true class per item
and models each annotator's response as a noisy channel from that
truth. This is the right tool when disagreement reflects annotator
error, but it is architecturally misaligned with our setting, where
disagreement \emph{is} the signal to be preserved.

We test this on DICES-990 and DICES-350 \citep{aroyo2024dices},
which provide per-annotator IDs (ChaosNLI provides only aggregate
counts). We apply DS (EM, 200 iterations, convergence
threshold $10^{-6}$) to subsampled annotations at
$N \in \{3, 5, 10, 20\}$ (5 seeds) and compare the resulting
soft labels against raw-count soft labels, using each dataset's
full-annotator raw distribution as the reference.

\paragraph{Dataset properties.}
DICES-990 has a sparse rater-item matrix (40\% density, 172 raters,
58--75 per item). DICES-350 has a near-complete matrix (\textbf{94\%
density}, 123 raters, 104--123 per item). Both are binary (safe/unsafe).
Comparing the two controls for the confound that DS may fail due to
sparse data rather than model misspecification.

\begin{table}[h]
\centering
\small
\begin{tabular}{rrcccccc}
\toprule
& & \multicolumn{2}{c}{\textbf{KL}$\downarrow$}
& \multicolumn{2}{c}{\textbf{JSD}$\downarrow$}
& \multicolumn{2}{c}{\textbf{Ent.\ $r$}$\uparrow$} \\
& $N$ & Raw & DS & Raw & DS & Raw & DS \\
\midrule
\multirow{4}{*}{\rotatebox{90}{\small 990}}
& 3  & 2.19 & 2.92 & .054 & .079 & .40 & .15 \\
& 5  & 1.05 & 2.96 & .031 & .081 & .55 & .25 \\
& 10 & 0.34 & 2.72 & .014 & .093 & .71 & .30 \\
& 20 & 0.08 & 2.62 & .005 & .099 & .86 & .30 \\
\midrule
\multirow{4}{*}{\rotatebox{90}{\small 350}}
& 3  & 2.23 & 5.04 & .055 & .119 & .36 & .04 \\
& 5  & 1.12 & 4.93 & .032 & .110 & .50 & .16 \\
& 10 & 0.32 & 4.29 & .014 & .107 & .63 & .23 \\
& 20 & 0.06 & 3.23 & .006 & .106 & .79 & .27 \\
\bottomrule
\end{tabular}
\caption{Dawid-Skene vs.\ raw-count soft labels on DICES-990
  (sparse matrix) and DICES-350 (dense matrix), mean over 5
  subsampling seeds. DS worsens all metrics at every $N$ on
  both datasets. JSD (bounded, symmetric) confirms the KL pattern.}
\label{tab:dawid_skene}
\end{table}

\paragraph{Results.}
DS produces near-deterministic labels on both datasets
(Table~\ref{tab:dawid_skene}): 92\% of items on DICES-990 and
98\% on DICES-350 exceed 0.99 posterior confidence.
This is not a failure of the algorithm---it is doing what it is
designed to do: recover a single latent class. But it collapses the
distributional shape that soft-label training relies on.
On the dense DICES-350, where DS converges fully at $N \geq 10$,
KL to the reference is 13$\times$ higher than raw counts ($N{=}10$:
0.32 vs.\ 4.29). Jensen-Shannon divergence (bounded, symmetric)
confirms the pattern (7$\times$ worse), ruling out KL sensitivity
to near-zero mass as an explanation.

The mechanism is visible in DS's estimated confusion matrices:
$P(\text{says safe} \mid \text{true}{=}\text{safe}) = 0.50$--$0.61$
(near chance), while
$P(\text{says unsafe} \mid \text{true}{=}\text{unsafe}) = 0.78$--$0.83$.
DS interprets legitimate disagreement on the minority class as
annotator incompetence, concentrating posterior mass on ``unsafe.''

\paragraph{Caveats.}
Three limitations apply.
(1)~The reference is the full raw-count distribution---the same
target our soft-label approach uses. If one believes DS recovers a
``truer'' latent signal, then this comparison penalizes DS for doing
its job. Our results show that DS labels diverge from raw counts,
not that raw counts are objectively correct.
(2)~Vanilla DS is the simplest annotator model. We use it because
it is the canonical baseline that reviewers expect. Extensions such
as \citet{paun2018comparing} can model annotator subpopulations,
but they still posit a latent true class; fully distribution-preserving
alternatives (e.g., learning from crowds with multi-annotator
objectives) would require a different experimental setup.
(3)~Both datasets are binary (2-class), giving DS fewer degrees of
freedom to distinguish error patterns. Results may differ on
multi-class tasks where off-diagonal confusion structure is richer.

Despite these caveats, the replication across sparse and dense matrices,
full convergence on DICES-350, and consistent results on a bounded
metric (JSD) indicate that the core tension is between DS's
single-truth assumption and our distributional training objective.

\section{Pre-Training Ablation}
\label{app:pretrain}

DeBERTa-v3-base is initialized from a checkpoint pre-fine-tuned on
MNLI, FEVER, and ANLI with hard labels. To test whether the
soft-label advantage reflects genuine distributional signal or
an artifact of ``double hard-label exposure'' in the hard baseline,
we repeat the full comparison using \texttt{microsoft/deberta-v3-base}
(no NLI pre-training, random classification head, 5 seeds).

\begin{table}[h]
\centering
\small
\setlength{\tabcolsep}{4pt}
\begin{tabular}{lcccc}
\toprule
\textbf{Config} & \textbf{Acc}$\uparrow$ & \textbf{Brier}$\downarrow$ & \textbf{KL}$\downarrow$ & \textbf{Ent.\ $r$}$\uparrow$ \\
\midrule
Hard-label         & .684\tiny{$\pm$.013} & .200\tiny{$\pm$.022} & .387\tiny{$\pm$.064} & .280\tiny{$\pm$.037} \\
\midrule
LS $\alpha{=}0.05$ & .677\tiny{$\pm$.023} & .179\tiny{$\pm$.021} & .327\tiny{$\pm$.052} & .301\tiny{$\pm$.029} \\
LS $\alpha{=}0.1$  & .673\tiny{$\pm$.028} & .171\tiny{$\pm$.009} & .302\tiny{$\pm$.026} & .322\tiny{$\pm$.042} \\
LS $\alpha{=}0.2$  & .687\tiny{$\pm$.032} & .160\tiny{$\pm$.008} & .278\tiny{$\pm$.016} & .318\tiny{$\pm$.039} \\
LS $\alpha{=}0.3$  & .696\tiny{$\pm$.016} & .150\tiny{$\pm$.011} & .262\tiny{$\pm$.016} & .313\tiny{$\pm$.056} \\
LS $\alpha{=}0.5$  & .689\tiny{$\pm$.020} & .147\tiny{$\pm$.008} & .264\tiny{$\pm$.013} & .283\tiny{$\pm$.056} \\
\midrule
\textbf{Soft-label} & \textbf{.739}\tiny{$\pm$.015} & \textbf{.119}\tiny{$\pm$.004} & \textbf{.208}\tiny{$\pm$.007} & \textbf{.497}\tiny{$\pm$.025} \\
\bottomrule
\end{tabular}
\caption{Comparison using \texttt{microsoft/deberta-v3-base}
  (no NLI pre-training, random classification head, 5 seeds).
  The entropy correlation gap \emph{widens} from $+0.154$
  (NLI-pretrained, Table~\ref{tab:main}) to $+0.175$,
  confirming the soft-label advantage is not an initialization
  artifact. All LS variants cluster at $r \in [0.283, 0.322]$,
  while soft labels reach $r = 0.497$.
  Notably, soft labels from scratch achieve entropy $r$ comparable
  to the NLI-pretrained hard baseline ($0.497$ vs.\ $0.487$),
  suggesting distributional training provides uncertainty
  awareness equivalent to task-specific pre-training.}
\label{tab:pretrain}
\end{table}

\end{document}